\documentclass[journal,10pt]{IEEEtran} 
\IEEEoverridecommandlockouts
\usepackage{lipsum}
\usepackage{amsmath,amssymb,amsfonts}
\usepackage{algorithmic}
\usepackage{graphicx}
\usepackage{textcomp}
\usepackage{xcolor}
\def\BibTeX{{\rm B\kern-.05em{\sc i\kern-.025em b}\kern-.08em
    T\kern-.1667em\lower.7ex\hbox{E}\kern-.125emX}}

\usepackage[english]{babel}
\usepackage[square,numbers]{natbib}
\usepackage{longtable}
\usepackage{etoolbox}
\usepackage{amsmath,amssymb,amsfonts}
\usepackage{algorithmic}
\usepackage{graphicx}
\usepackage{textcomp}
\usepackage{xcolor}
\usepackage{natbib}
\usepackage[hyphenbreaks]{breakurl}
\usepackage[hyphens]{url}

\usepackage[utf8]{inputenc} 
\usepackage[T1]{fontenc}    
\usepackage{hyperref}       
\usepackage{url}            
\usepackage{booktabs}       
\usepackage{amsfonts}       
\usepackage{nicefrac}       
\usepackage{microtype}      

\usepackage{multirow}
\usepackage{makecell}
\usepackage{amsmath}        
\usepackage{amssymb}        
\usepackage{amsthm}
\usepackage{tikz}
\usepackage{pdfpages}
\usepackage{comment}
\usepackage{flushend}
\usepackage[linesnumbered,ruled]{algorithm2e}
\usepackage[justification=centering]{caption}

\usepackage{fancyvrb}
\usepackage{datetime}
\newdate{date}{25}{01}{2023}
\date{\displaydate{date}}

\newenvironment{noindlist}
 {\begin{list}{\labelitemi}{\leftmargin=0.3em \itemindent=0.6em}}
 {\end{list}}

 \newenvironment{noindenumerate}
  {\begin{list}{\arabic{enumi}.}
    {\usecounter{enumi}
     \setlength{\labelwidth}{0.6cm}
     \setlength{\leftmargin}{0.33cm}
     \setlength{\labelsep}{0.1cm}
     \setlength{\itemindent}{0cm}
     }
  }
  {\end{list}}
\begin{document}
\pagenumbering{gobble} 
\pagestyle{empty}
\title{Prompting Large Language Models \\ With the Socratic Method}

\author{Edward Y. Chang \\ Computer Science, Stanford University \\ echang@cs.stanford.edu%
}


\maketitle


\pagestyle{empty}
\begin{abstract}
This paper presents a systematic approach to using the Socratic method in developing prompt templates that effectively interact with large language models, including GPT-3. Various methods are examined, and those that yield precise answers and justifications while fostering creativity and imagination to enhance creative writing are identified. Techniques such as {\em definition}, {\em elenchus}, {\em dialectic}, {\em maieutics}, {\em generalization}, and {\em counterfactual reasoning} are discussed for their application in engineering prompt templates and their connections to inductive, deductive, and abductive reasoning. Through examples, the effectiveness of these dialogue and reasoning methods is demonstrated. An interesting observation is made that when the task's goal and user intent are conveyed to GPT-3 via ChatGPT before the start of a dialogue, the large language model seems to connect to the external context expressed in the intent and perform more effectively.
\end{abstract}


\section{Introduction}
\label{sec:intro}

Prompting is a technique used to guide the output generation of a pre-trained language model such as GPT-3 \cite{OpenAI-GPT3-2020}. This is achieved by providing input in the form of a question or template, which helps to generate specific responses such as Q\&A, document summarization, and translations. The advent of ChatGPT \cite{ChatGPTvsHuman, chatgpt, wolf2019transfertransfo} has revolutionized the field of NLP by demonstrating the potential of using large pre-trained language models with prompting. Despite this progress, there is still room for improvement in current prompting strategies and techniques, especially for specific target applications. In this study, we investigate the Socratic method \cite{PaltoRepublicURL,SocraticMethidWiki} to identify and evaluate potential prompting strategies, and use the findings to design effective prompt templates.

Traditional NLP tasks involve various sub-tasks, such as named entity recognition, dependency parsing, coreference resolution \cite{dobrovolskii-2021-word}, semantic parsing \cite{pasupat-liang-2015-compositional,dong-lapata-2018-coarse}, and more, to comprehend the meaning of a sentence. By utilizing prompt templates with large language models (LLMs), these sub-tasks can be delegated to the LLM, freeing the template to focus specifically on dialogue design. In this regard, the Socratic method \cite{PaltoRepublic} holds significant relevance, as it is well-known for using questioning (prompting) as a means of promoting critical thinking and delving into complex concepts \cite{Elder2010}.

The Socratic method has a long history of being regarded as the basis of critical thinking. However, some recent studies have cast doubt on its effectiveness in practice. In his paper ``Socratic Irony and Argumentation,'' Airaksinen \cite{Irony2022} criticizes the method for its rigidly defined roles of teacher and student, which can lead to fear of not meeting the teacher's expectations and reluctance to participate. Similarly, Stoddard's ``The Use of Socratic Questioning in Clinical Teaching'' \cite{PimpClinical2016} highlights the risk of the method being misused in a manner that lacks psychological safety for students. Fortunately, when using the Socratic method in a dialogue with an LLM, the absence of emotions and sarcasm, as well as the option to deactivate the model, can alleviate many of the problems associated with human interaction.

This study starts by presenting an overview of the Socratic method's strategies and techniques. To begin, we list ten widely referenced methods \cite{AskRightQ2001} under the Socratic method umbrella and use hypothesis elimination to identify the most relevant ones for our goal of prompt-template development. The selected methods are definition, hypothesis elimination, elenchus, dialectic, maieutics, generalization, and induction. Furthermore, we add to the list counterfactual reasoning, which is a concept in logic that involves considering what might have happened if a particular event had occurred differently. We then perform experiments using GPT-3 to test and evaluate these methods, and offer suggestions for incorporating these strategies and techniques into prompt templates.

In their work on ``Critical Thinking: The Art of Socratic Questioning,'' Paul and Elder identify three types of Socratic questioning: spontaneous, exploratory, and focused \cite{Paul2007CriticalTT}. We will not discuss spontaneous questioning, as it is similar to casual conversation. Focused questioning (type 2), on the other hand, is geared towards gaining knowledge and truth, and methods such as {\em definition}, {\em elenchus} (cross-examination), {\em hypothesis elimination}, {\em dialectic}, and {\em generalization} hold great potential for developing effective prompting strategies and improving the response accuracy of a large language model (LLM). An interesting observation is that when the user intent is conveyed to GPT-3 during the task {\em definition} stage, before the start of a dialogue, the LLM seems to connect to the external context expressed in the intent and perform more effectively. (Table~\ref{tab:genesis} provides an example of pre-dialogue warm-up. More examples are documented in \cite{CRITExtended2023}.)

Additionally, exploratory thinking (type 3) can be supported through the {\em maieutics} (midwife) method, {\em induction}, and {\em counterfactual reasoning}, which can guide GPT-3 towards producing imaginative and creative writing. While many of the plot suggestions generated by GPT-3's exploration may not be useful, a few unique recommendations in response to a "what if" query can stimulate the writer's imagination and lead to remarkable results. When applied effectively, these methods can turn an LLM into a writer's muse, providing inspiration and guiding the creative process \cite{MuseMasses2010}.

The main contributions of this paper are as follows:

\begin{noindlist}
\item An overview of the Socratic method's strategies, their evaluation, and selection of the most relevant ones for the development of effective prompt templates.
\item An examination of how the definition, elenchus, hypothesis elimination, dialectic, and generalization methods can improve the output's accuracy and conciseness through clarification and verification.
\item An illustration of how maieutics, induction, and counterfactual reasoning can foster productive generalization and creativity.
\end{noindlist}

The remainder of this paper is structured into five sections. Section~\ref{sec:related} provides a review of related work on prompting methods in natural language processing. In Section~\ref{sec:socratic}, we introduce the ten strategies and methods taught by Socrates and used in Plato's ``Dialogues.'' From these, we select relevant methods along with counterfactual reasoning as our focus for developing prompting templates. Section~\ref{sec:template} details how we engineer these methods into our templates to improve output correctness and stimulate creative writing. In Section~\ref{sec:pilot}, we present a pilot study. Finally, in Section~\ref{sec:conc}, we present our concluding remarks.

\section{Related Work}
\label{sec:related}

The use of transformer architecture \cite{Vaswani2017AttentionIA} and masked data for pre-training large language models (LLMs) in an unsupervised setting has become {\em the approach} in natural language processing \cite{Devlin2019BERTPO,Lewis2019BARTDS}. The method involves pre-training an LLM on a large text corpus, followed by fine-tuning for specific tasks.

Prompting is a recent innovation in the field, popularized by OpenAI, especially with the release of GPT-3 in 2020. Instead of fine-tuning the model for a specific task, the approach involves providing a specific input, or ``prompt,'' to guide the LLM's output generation, resulting in greater flexibility and efficiency in generating a wide range of responses.

However, designing effective prompt templates remains a challenge \cite{METASurvey2023}, as it requires a deep understanding of the interplay between the LLM and the prompt. According to the survey paper \cite{SocraticModels-Google2022}, there are several factors that impact prompt template engineering, including the type of LLM used, manual vs automatic design, and static vs continuous prompts.

\begin{noindlist}
\item {Left-to-right vs masked LLMs}. For tasks related to generation or tasks solved using a standard left-to-right language model \cite{OpenAI-GPT3-2020}, prefix prompts tend to perform better, as they align with the model's left-to-right nature. For tasks solved using masked language models \cite{Devlin2019BERTPO}, cloze prompts are more suitable, as they closely match the pre-training task form.
\item {Manual vs automatic design}. A prompt template should be tailored to the specific LLM. While manual design may be suitable in the initial flow-design phase, dependencies between the input and expected output, and their variations, should be mined automatically \cite{WhatLMKnowsJiang2020}. Automation can also help in paraphrasing the seed prompt to support various mined dependency patterns, but mistakes can occur \cite{PTRSUN2021}.
\item {Discrete vs continuous prompts}. Discrete prompts involve providing a fixed set of pre-determined input choices to an LLM. Continuous prompts, on the other hand, involve a dialogue or conversation between the model and the user, allowing for a more dynamic and interactive experience. 
\end{noindlist}

More advanced templates can be constructed by combining basic templates with techniques like ensemble methods \cite{Schick2020ExploitingCF}. This involves forming a committee of basic templates that ask the same question using different phrasing \cite{Haviv2021BERTeseLT}.
Most current prompt templates generate short outputs, such as class labels, or outputs with a length that can be predicted based on the task and input, like in the case of translation. However, for tasks that may generate longer or open-ended outputs, additional considerations may be necessary during the template engineering process. 

One approach for generating longer outputs is explanation-based prompting, as proposed by the chain-of-thought method \cite{wei2022chain}. This method generates a sequence of explanations before inferring the answer. However, when dealing with simple math problems, this approach has an error rate of $47\%$. To address the inconsistency issues of explanation-based prompting, \cite{Jung2022MaieuticPL} formulates the problem as a satisfiability problem, which defers inference until a tree of explanations has been expanded abductively (explaining both truth and false branches) and recursively. However, using abductive reasoning alone is often considered weak, incoherent, and even nonexistent \cite{ReasoningLLM2022,Abductive4Flaws2011}. To improve consistency, a recent work \cite{wang2023selfconsistency} extends the chain-of-thought approach by adding a diverse set of reasoning paths and performing majority voting among them. This method can be viewed as an ensemble method, but it does not alter the nature of abductive reasoning.

In contrast, the Socratic method aims to employ deductive, inductive, and abductive reasoning to ensure consistency and accuracy of inference. The Socratic method deals with all aspects of critical thinking, including definition clarification and cross-examination. This comprehensive approach to template engineering can lead to improved output quality and consistency. 


The primary objective of this study is to design continuous prompts that enhance response quality and foster guided creativity in generative tasks, such as verifying information, evaluating source credibility, proposing alternatives, recommending plot ideas in creative writing, and generating task-specific surprises. Our approach involves investigating strategies and methods within the Socratic method, and selecting the most relevant ones for further exploration.

As discussed in Section~\ref{sec:intro}, Socratic questioning can be classified into three categories: spontaneous, exploratory, and focused \cite{Paul2007CriticalTT}. When designing a prompt, it is important to consider the category and utilize the most suitable strategies and techniques to achieve the best results. 
\section{The Socratic method}
\label{sec:socratic}

The Socratic method is a questioning technique used in teaching and philosophy to encourage critical thinking and self-discovery \cite{SocraticMethidWiki}. The method involves asking a series of questions to explore complex ideas and help individuals arrive at their own understanding of a concept. It is based on the belief that knowledge cannot be simply imparted, but must be discovered through a process of questioning and dialogue.

Some of the Socratic method's key principles and guidelines to conduct critical thinking include:
\begin{noindlist}
\item Posing open-ended questions: The teacher or facilitator starts with a question to stimulate thinking and draw out ideas.
\item Clarifying key terms: The teacher helps the students clarify and define relevant terms and concepts to ensure everyone is on the same page.
\item Providing examples and evidence: The teacher or facilitator encourages the students to provide examples and evidence as reasons to support their claims.
\item Challenging reason-to-conclusion argument: The teacher or facilitator challenges the students' arguments and encourages them to question their own beliefs and to consider alternative perspectives.
\item Summarizing and drawing conclusions: The teacher helps the students summarize and draw conclusions from the discussion.
\item Reflecting on the process: The teacher and students reflect on the effectiveness of the method and what they learned through the dialogue.
\end{noindlist}

These principles of the Socratic method are realized through various methods and strategies. (Note the term ``method'' are used at the abstract level referring to the Socratic teaching
through questioning method, and his specific questioning techniques.) Some well-known examples of the Socratic method in action include Plato's ``Dialogues'' and ``Republic'' \cite{PaltoRepublicURL}, where Socrates uses questioning to  explore complex ideas and stimulate critical thinking in his interlocutors.

\begin{noindenumerate}
\item Definition: Socrates is known for his use of definition to clarify and explain the meaning of key terms and concepts.
\item Generalization: This method draws general principles from patterns that underlie observations and theories.
Generalization is used to form more certain and comprehensive conclusions.

\item Induction: Similar to generalization, but induction is based only on empirical evidence. Inductive reasoning provides hypotheses with high uncertainty.

\item Elenchus: This method involves cross-examination, where a series of questions is used to test the consistency and coherence of hypotheses and beliefs. Elenchus aims to 
test the validity of someone's arguments and to help them refine their thinking and eventually come up with well-supported hypotheses.

\item Hypothesis Elimination: This method involves eliminating false hypotheses and beliefs by testing them against counterexamples and logical reasoning.  Different from method elenchus, hypothesis elimination tests a hypothesis against evidence and logic to determine if it is true or false. 

\item Maieutics: This method involves helping individuals bring out the knowledge and understanding they already possess.
Maieutics is conducted by asking questions that encourage the person to reflect on their own experience, knowledge, beliefs and to explore alternative perspectives. Maieutics fosters  self-discovery, creative writing, and innovation.

\item Dialectic: This method involves exploring opposing viewpoints through dialogue or debate to arrive at a deeper understanding of a subject.

\item Recollection: This method involves the belief that knowledge is innate, and that people can remember what they already know through a process of questioning.

\item Irony: This method involves exposing ignorance and pretensions through irony, and pointing out the gap between claims and true understanding.

\item Analogy: This method involves comparing and contrasting different concepts through analogies, in order to help individuals understand complex ideas.
\end{noindenumerate}


At first glance, some reasoning methods may seem similar. For example, both induction and generalization use inductive reasoning, while both elenchus and hypothesis elimination use deductive reasoning. Similarly, methods like definition and dialectic use both inductive and deductive reasoning to explore opposing viewpoints through dialogue or debate. However, it is important to note that these methods have distinct differences, which will be discussed later in this paper.

In the context of critical thinking, methods like definition, elenchus, dialectic, hypothesis elimination, and generalization play active roles. On the other hand, during the brainstorming stage or in the context of creative thinking, methods like maieutics, induction, and counterfactual thinking are more relevant.

Analogy, irony, and recollection, are less relevant to our goal, so we do not consider them. Irony and analogy may not be necessary when working with language models, as these models may not understand figurative language. Recollection is limited by the memory of ChatGPT and GPT-3, which is a context window of $4k$ and $8k$, respectively. The prompter must use this limited space as context to allow the language model to recall information.

\subsection{Illustrative Critical Reading Example}

To illustrate how these methods can practically be applied, let's use the example of critical reading. Critical reading is a crucial component of critical thinking, which involves evaluating the quality and credibility of written materials, from research papers to blog posts \cite{lai-etal-2017-race,PaulBinkerCT1990}. It requires a systematic and analytical approach, asking relevant questions, and using effective prompts to gain deeper understanding of the text \cite{Elder2010}.

\begin{table}[ht!]
\begin{center}
\begin{tikzpicture}
\node (table) [inner sep=0pt] {
\begin{tabular}{|p{0.56cm}|p{6.9cm}|}
\toprule
\textbf{} & \textbf{Function $\Gamma$ = CRIT($d$)} \\
\midrule
& \textbf{Input}. $d$: document; \textbf{Output}. $\Gamma$: validation score; \\
& \textbf{Vars}. $\Omega$: claim; $R$ \& $R'$: reason \& counter reason set; \\
& \textbf{Subroutines}. $Claim$(), $FindDoc$(), $Validate$(); \\
& \textbf{Begin} \\
\#1 & {\hspace{.2cm}}Identify in $d$ the claim statement $\Omega$; \\
\#2 & {\hspace{.2cm}}Find a set of supporting reasons $R$ to $\Omega$; \\
\#3 & {\hspace{.2cm}}For $r \in R$ eval $r \Rightarrow \Omega$ \\
& {\hspace{.5cm}}{If} $Claim$($r$), ($\gamma_r$, $\theta_r$) = CRIT($FindDoc$($r$)); \\
& {\hspace{.5cm}}{else}, ($\gamma_r$, $\theta_r$) = $V$($r \Rightarrow \Omega$); \\
\#4 & {\hspace{.2cm}}Find a set of rival reasons $R'$ to $\Omega$; \\
\#5 & {\hspace{.2cm}}For $r' \in R'$, ($\gamma_{r'}$, $\theta_{r'}$) = V($r' \Rightarrow \Omega$) eval rival arguments; \\
\#6 & {\hspace{.2cm}}Compute weighted sum $\Gamma$, with $\gamma_r$, $\theta_r$, $\gamma_{r'}$, $\theta_{r'}$. \\
\#7 & {\hspace{.2cm}}Analyze the arguments to arrive at the $\Gamma$ score. \\
\#8 & {\hspace{.2cm}}Reflect on and synthesize CRIT in other contexts. \\
& \textbf{End} \\
\bottomrule
\end{tabular}
};
\draw [rounded corners=.5em] (table.north west) rectangle (table.south east);
\end{tikzpicture}
\end{center}
\caption{CRIT Pseudo-code \cite{CRITExtended2023}. (The symbol $\Rightarrow$ denotes both inductive and deductive reasoning.)}
\label{tab:CRIT}
\end{table}

To aid in critical reading, we introduce a template called CRIT \cite{CRITExtended2023}, which stands for Critical Reading Inquisitive Template\footnote{It is important to note that the CRIT template presented here is intended for analyzing research, opinion, and news articles, and is not suitable for analyzing literature such as novels, prose, or poetry. Each type of literary work has its unique style and nature, which require tailored prompts to facilitate effective analysis.}. Given a document $d$, CRIT evaluates it and produces a validation score $\Gamma$. Let $\Omega$ denote the conclusion or claim of $d$, and let $R$ be the set of reasons supporting the claim. We define ($\gamma_r, \theta_r$) = V($r \Rightarrow \Omega$) as the causal validation function, where $\gamma_r$ denotes the validation score, $\theta_r$ the source credibility score,  for each reason-to-conclusion argument $r \Rightarrow \Omega$. Table~\ref{tab:CRIT} presents the pseudo-code of $\Gamma$ = CRIT($d$), which generates the final validation score $\Gamma$ for document $d$ with justifications.

In the following subsections, we will discuss how CRIT uses these five methods: 1) definition, 2) elenchus, 3) dialectic, 4) maieutics, and 5) counterfactual thinking.

\subsection{Method of Definition}

As shown in the pseudocode in Table~\ref{tab:CRIT}, 
the CRIT algorithm starts in
its step $\#1$, asking GPT-3 to identify the conclusion of a document. To avoid any misunderstandings, the prompt includes a clear instruction and definition. (In the square brackets, symbol {\em in} denotes a input slot to an LLM and {\em out} the output slot.)

\begin{table}[ht!]
\small
\begin{tabular}{p{0.6cm}|p{7.2cm}}
p1.1 & ``What is the conclusion in document [in: $d$] 
  [out: $\Omega$]? \\ 
& The conclusion statement may 
  be written in the last paragraph, near 
  keywords "in conclusion," "in summary," or "therefore."'' \\
\end{tabular}
\end{table}

We can use
the {\em definition} method to 
improve the understanding of the document. 
One approach is paraphrasing the prompt 
into multiple prompts and grouping them into an ensemble, similar to forming a thesis committee. (Section~\ref{sec:template} presents
prompt ensemble in details.)
Different members can phrase the same question in different ways or ask it from a different perspective.  For example:

\begin{table}[h!]
\small
\begin{tabular}{p{0.6cm}|p{7.2cm}}
p1.2 &  ``What is the issue addressed by [in: $d$] 
  [out: $\Omega$]?'' \\
p1.3 & ``What is the most important outcome presented in text [in: $d$]? [out: $\Omega$]''
\end{tabular}
\end{table}

Step $\#2$ in Table~\ref{tab:CRIT} prompts GPT-3 to find a set of supporting reasons. To further enhance the accuracy and comprehensiveness of the results, the prompt can ask for not only ``reasons'' but also ``theories,''``evidences,'' or ``opinions'' to query for the document's support to its conclusion, similar to the ensemble method.

\begin{table}[ht!]
\small
\begin{tabular}{p{0.6cm}|p{7.2cm}}
p2 & ``What are the supporting reasons [out: $R$] of conclusion \\
& [in: $\Omega$] of [in: $d$]? A reason can be a theory evidence or opinion.'' 
\end{tabular}
\end{table}

\subsection{Method of Elenchus}

The method of elenchus is rooted in the Greek word ``elenchein,'' which translates to examine. This method involves cross-examining the results generated by GPT-3 to evaluate the consistency and coherence of the arguments. The goal is to arrive at a deeper understanding of the validity of the reasons and conclusion, and to identify any potential weaknesses or flaws in the arguments.

Step $\#3$ of the CRIT algorithm prompts GPT-3 to assess the validity of each reason $r \in R$ as justification for the conclusion $\Omega$ through the function V($r \Rightarrow \Omega$). To validate the reason-to-conclusion argument, CRIT must evaluate the presented reason and its causal relationship with the conclusion and conduct cross examination, which is precisely the task of the method of elenchus.

CRIT issues four prompts in step $\#3$ to evaluate the logic validity and source credibility of the $r \Rightarrow \Omega$ reasoning. CRIT first elicits supporting evidence for reason $r \in R$. This evidence can be a theory, an opinion, statistics, or a claim obtained from other sources. If the reason itself is a claim, then the sources that the claim is based on are recursively examined. The strength of the argument and its source credibility are rated on a scale of $1$ to $10$, with $10$ being the strongest.

\begin{table}[h!]
\small
\begin{tabular}{p{0.6cm}|p{7.2cm}}
p3.1 & ``What is the evidence for reason [in: $r$] to support
conclusion [in: $\Omega$] in document [in: $d$]? [out: evidence]'' \\ 
p3.2 & ``What is the type of evidence? A) a theory, B) an opinion, C) statistics, or {\color{red}D}) a claim from other sources?'' \\
p3.3 & ``If the evidence of reason [in: $r$] is {\color{red}D}),
  call CRIT recursively'' \\
p3.4 & ``How strongly does reason [in: $r$] support
  [in: $\Omega$] in document [in: $d$]? Rate argument validity [out: $\gamma_r$] and source credibility [out: $\theta_r$] between $1$ and $10$ (strongest).''
\end{tabular}
\end{table}

It may be beneficial to also incorporate the counter-argument method in order to gain a more comprehensive and balanced evaluation of the argument. This can result in a deeper understanding of the topic being discussed. We will be discussing this further in the next section.

\subsection{Method of Dialectic}

The easiest way to mislead without lying outright is to leave out critical counterarguments from the reader.
CRIT relies on GPT-3 to generate and
evaluate counter arguments, similar to
how it prompts GPT-3 to extract and 
evaluate reasons. 

CRIT in its step $\#4$ asks GPT-3 to provide missing
rival reasons, and then pair rival reasons
with the conclusion to conduct validation.
There are two strategies to 
bring counter arguments to the surface.
The first strategy attacks the weakest 
arguments with the lowest scores and asking
GPT-3 to attack those arguments.

\begin{table}[h!]
\small
\begin{tabular}{p{0.6cm}|p{7.2cm}}
p4 & ``Is there a counterargument against [in: $r \Rightarrow \Omega$]? If so, provide counter reasons [output $R'$].'' \\
p5 & Similar to p3, except for replacing argument $r$ with rival argument $r'$.
\end{tabular}
\end{table}

For finding omitted information, CRIT can
query GPT-3 without quoting any $r \in R$,
and follow the same process. 

Next, in step $\#6$, CRIT computes an aggregated score by performing a weighted sum on the validation multiplied by the credibility scores of both arguments and counterarguments, and then outputs the final assessment score $\Gamma$.

\begin{table}[h!]
\small
\begin{tabular}{p{0.6cm}|p{7.2cm}}
p6 & ``Final score [out: $\Gamma$]. $\Gamma = 
\sum_{r \in R \cup R'} \gamma_r \times \theta_r / |R \cup R'|$.
\end{tabular}
\end{table}

\subsection{Method of Maieutics}
\label{sec:Maieutics}

The maieutic method derives from the Greek word ``maieutikos,'' meaning midwife. It is founded on the belief that a teacher's role is to facilitate students in bringing forth their own understanding of a subject, rather than simply conveying knowledge. Unlike the elenctic method, which aims to detect and eliminate false hypotheses, maieutics centers on helping students reveal their own understanding of a subject. In this dialogical method, the teacher asks questions that are intended to guide the student in discovering their own comprehension, rather than providing them with information or answers.

Continuing with GRIT, once the text has been scored in step $\#6$, it can be valuable for readers or students to enhance their analytical and writing skills by summarizing and analyzing the justifications produced by GPT-3. CRIT in its step $\#7$ can prompt GPT-3 to generate a report, which readers and students can then compare with their own notes.

\begin{table}[h!]
\small
\begin{tabular}{p{0.6cm}|p{7.2cm}}
p7 & ``For every $r \in R \cup R'$ justify the validity score $\gamma_r$ and source credibility score $\theta_r$ for argument $r \Rightarrow \Omega$.''
\end{tabular}
\end{table}

\subsection{Counterfactual Reasoning}
\label{sec:cf}

Counterfactual reasoning~\cite{WinArgument2006,Cross-Examination2021} can be seen as a natural extension of the Socratic method, as both involve questioning assumptions and exploring alternative perspectives. 
Counterfactual thinking involves imagining alternative scenarios to what actually happened, often using phrases like ``what if'' or ``if only.''
By incorporating counterfactual reasoning into prompt engineering, one can facilitate exploration of alternative possibilities and promote more nuanced and complex understanding of a given topic.

The final step of GRIT involves using the counterfactual method to encourage students to reconsider the arguments and counterarguments presented in the text based on new contextual information. CRIT can prompt students with questions such as ``what if the debate in the text took place now instead of in the 1950s?'' or ``what if the main event in the text occurred in Asia instead of in Europe?'' Students can express their own opinions and findings based on further reading and statistics, and challenge the conclusions drawn in the text. 

\begin{table}[h!]
\small
\begin{tabular}{p{0.6cm}|p{7.2cm}}
p8 & ``For every $r \in R \cup R'$, evaluate $r \Rightarrow \Omega$
in [in context].''
\end{tabular}
\end{table}

\subsection{Remarks on the Socratic Method and CRIT}

As we have shown that for critical reading,
GRIT uses three methods, definition, elenchus, and dialectic.  
For critical thinking, CRIT uses methods maieutics and counterfactual reasoning. For more explorative 
thinking, methods such as induction can be used for informal brainstorming, hypothesis elimination for removing weak propositions, and generalization for deriving principles from examples.

Please note that prompts can be submitted to GPT-3 either all together or one-by-one. Our empirical study on reading comprehension samples~\cite{501Q2004} demonstrates that issuing prompts one-by-one results in outputs with finer details. This is because GPT-3 has the opportunity to analyze a document multiple times for slightly different purposes. For teaching critical reading to K-12 students, one-by-one prompting is preferred as it allows students to engage with CRIT step-by-step.  However, for answering multiple-choice
questions, both prompting all together and one-by-one receive 
similar scores. We will conduct large-scale study with
ablation tests to investigate if adding or
deleting prompts and using different submission methods
make marked differences.

\section{Prompt Template Engineering}
\label{sec:template}

Prompt template engineering involves creating templates to provide input, or ``prompts,'' to a language model to guide its output generation. In this section, we discuss prompt template engineering methods for basic building blocks, and then integrate the methods of definition, elenchus, dialectic, maieutics, and counterfactual reasoning to compose more complex templates. We present experimental results using different types of documents to demonstrate how the Socratic method can improve the accuracy and conciseness of the output through arguments and verification, as well as facilitate guided generalization and creativity.

\subsection{Basic, One Shot Template}

Let's begin by discussing a simple one-shot prompt template. In the work of \cite{SocraticModels-Google2022}, a simple formulation function is used to generate the prompt $x'$, which is obtained by applying the function $f_{prompt}(x)$ to the input $x$.

For machine translation, the prompt template can take the form of ``Translate from [Lan$_{from}$]: [X] to [Lan$_{to}$]: [Y],'' where Lan$_{from}$ can be either detected by the prompt template or identified by the LLM. The input $x$ provides the information to fill in the slots [X] and [Lan$_{to}$]. For example, if the input is ``translate good morning to French,'' the prompt template $x'$ would be ``Translate from English: 'good morning' to French: [Y].'' The empty slot [Y] is then filled with the LLM's output, such as ``bonjour.'' In cases where the LLM produces multiple responses, it can also provide a score for each, which the prompt template can use to select the highest-scoring response or to request a summary from the LLM.

There are three main design considerations when engineering a basic prompt. 
\begin{noindenumerate}
\item {Input style}. It is important to consider how to phrase the template so that it can handle different styles of user input for the same task. For example, a user may ask for a translation task to be performed by saying ``Translate $x$ to French,'' or ``What is the French translation of $x$?''
\item {LLM capability}. As discussed in \cite{PromptSurvey2023}, it is important to take into account the patterns and capabilities of the partner language model (LLM) when designing the template, such as whether the LLM is left-to-right \cite{OpenAI-GPT3-2020} or masked \cite{Devlin2019BERTPO}.
\item {Cost}. Certain tasks, such as language detection and summarization, can be performed by the template itself or by the LLM. The decision of whether to perform a task within the prompt template or to use the LLM should be based on factors such as cost.
\end{noindenumerate}

To address the first two technical challenges, one can start by hand-engineering a few seed templates and then paraphrasing them into an ensemble \cite{Haviv2021BERTeseLT}. We believe that the basic, one-shot formulation can always be replaced by an ensemble formulation \cite{Peng2022ModelEI, Schick2020ExploitingCF} and then learn the weights of its members for each query instance to produce the final output. Additionally, by examining which basic prompts have high weights, an ensemble with various paraphrased prompts can identify what an LLM knows, which can help infer its strengths without having to conduct capability
mining on the LLMs.


\subsection{Prompt Clarification with Method Definition}
\label{sec:template-definition}

There are computer algorithms that can already be used to recursively clarify a question, its definitions, and sub-terms' definitions. In fact, the natural language processing (NLP) community has developed a large 
number of useful methods and algorithms over the years \cite{NLP-Text-JM3}. 
One can use NLP techniques, such as dependency parsing and named-entity recognition (NER) \cite{NLPScratch2011}, to analyze the structure and meaning of a question and identify key terms and concepts. For example, 
NER can be used to extract entities in user input, such as names, locations, and organizations, and co-reference resolution can be used to understand the referred entity of a pronoun. 
Before submitting a template to an LLM, the application (e.g., a chatbot) 
that uses the template should check if all input slots are filled, 
and perform a sanity check. In the translation example, if the [Lan$_{to}$] was not provided or the specified language is not supported by the LLM, then the application should inquire the user for clarification.

Regarding mapping a natural language input to a prompt template, existing techniques of knowledge representation and reasoning can be very helpful. More specifically, ontology alignment and semantic parsing \cite{Campagna2020AFS,zhou-etal-2021-structure} can help map an NL input to a structured representation of knowledge and infer implicit concepts and relationships. These algorithms can be used to generate more precise and accurate prompts for LLMs, and to improve the effectiveness of the Socratic method in dialogue formulation \cite{DialoguewithAttention2023}. Some available tools include NLTK (Natural Language Toolkit) and spaCy for NLP, and TensorFlow for ML.


\subsection{Prompt Verification with Method Elenchus}

The main purposes of conducting cross examination in a template are to validate the credibility of the information sources and to identify inconsistencies in the process. Cross examination is typically conducted through a multi-turn dialogue \cite{DialoguewithAttention2023}. In the context of template engineering, the goal is to formulate a productive dialogue that can be used to assess the reliability of an LLM's output.  

There are several methods that can be used to assess and strengthen the reliability of an LLM's output. 1) The first approach is to paraphrase a question in order to obtain different answers and identify inconsistencies, if they exist, in multiple answers. 2) The second method is to ask for further evidence, such as querying top-k sources of information and asking the LLM to rate the credibility of each source. This can be used to compute the reliability of the output. 3) Additionally, template engineering can be used to query an LLM for opposing views of its output, including sources and credibility, and then evaluate if a different perspective is strong.

The implementation of the first two methods for cross examination, paraphrasing a question and asking for further evidence, is readily covered by the techniques enumerated in Section~\ref{sec:template-definition}. To implement the third method of asking for different perspectives, a simple approach is to find the sentiment of the original question and then rewrite the question with an opposite sentiment. For example, if the original question is phrased in a positive tone, the prompt template can reformulate the question with a negative tone to elicit a contrasting viewpoint.
A more elaborate method is to identify the people and sources in the LLM-generated responses and then re-post the questions to those who have a reputation for having different views. For example, if the original answer came from a democratic right-leaning source, the prompt template may post the same question to a source of a republican-left persuasion, and vice versa. This approach allows for a more comprehensive examination of the topic by considering multiple perspectives.

\begin{table}[t!]
\begin{tikzpicture}
\begin{small}
\node (table) [inner sep=0.1pt] {
\begin{tabular}{p{1.0cm}|p{6.6cm}}
\toprule
Role & Dialogue \\
\midrule
Prompt & {The discovery of the double helix structure of DNA was credited to James Watson and Francis Crick. However, are there opposing opinions to the list of contributors regarding this discovery?}\\

GPT3 & {There are some historical debates on the contributions of other scientists who also worked on the structure of DNA such as Maurice Wilkins and Rosalind Franklin.} \\ 
\midrule 

Prompt & 
{What is the tallest building in the world.  Please provide three references and rate their credibility as high, medium, or low.} \\

GPT3 & {As of 2021, the tallest building in the world is the Burj Khalifa located in Dubai, United Arab Emirates. It stands at 828 meters (2,722 feet) tall.} \\


& {1. Credibility {\color{red}High}: ``Burj Khalifa.''... 
 this is the official website of the building, providing accurate information.} \\

& {2. Credibility {\color{red}High}: ``The Tallest Buildings in the World.'' Council on Tall Buildings and Urban Habitat... this is a well-respected organization in the field of tall building and urban habitat research.} \\

& {3. Credibility {\color{red}Medium}: ``The Tallest Buildings in the World.'' World Atlas..., 
this is a well-known website providing accurate information on different topics.} \\
\bottomrule
\end{tabular}
};
\draw [rounded corners=.5em] (table.north west) rectangle (table.south east);
\end{small}
\end{tikzpicture}
\caption{Examples of Method Elenchus.}
\label{tab:Elenchus}
\vspace{-.25in}
\end{table}

The template to examine the semantic relation between two sentences $S_1$ and $S_2$ can be 
written as ``<$S_1$>, [R], [$S_2$],'' where R
is one of the three most important 
types of semantic relations: paraphrase, entailment, and contradiction \cite{PTRSUN2021}. Two sentences that have the same meaning are called paraphrases of each other. Two sentences that
have different meanings can be called 
disagreement or contradiction. The template 
can be trained to identify the degree
of agreement (or disagreement) between two
sentences.

Table \ref{tab:Elenchus} shows two examples of this. In the first example (shown on the top portion of the table), the prompter asks GPT-3 to confirm if James Watson and Francis Crick are the only contributors to the discovery of the DNA double helix structure. GPT-3 replies by mentioning two other contributors. The second example in the table asks GPT-3 to provide not only the answer to a question but also its information sources and rate the credibility of each source according to the prompter's specification. Although the reliability of GPT-3's ratings remains to be validated\footnote{Credibility of a source can
be evaluated based on an algorithm similar to Google's PageRank \cite{PageRank1998}.}, this rating mechanism can serve as an alert when some sources are found to be unreliable.

\subsection{Prompt Generalization with Method Maieutics}

\begin{table}[htbp]
\begin{tikzpicture}
\begin{small}
\node (table) [inner sep=0pt] {
\begin{tabular}{p{1.0cm}|p{6.8cm}}
\toprule
Role & Dialogue \\
\midrule
Prompt & {Can you interpret the meaning of this Taiwanese idiom: "planting gourd yields cucumber?"}\\
GPT3 & {The idiom is often used to convey the idea that sometimes, despite your best efforts, things don't turn out the way you want them to.} \\
Prompt & {You are half right. It also implies that the price of gourd is much higher than that of cucumber. The farmer not only yields unexpected results, but also loses income! } \\

GPT3 & {It's an idiomatic expression that can be used to describe a situation in which someone works hard and invests a lot of resources into something, but the outcome is not only different from what was expected, but also less profitable. It conveys the idea of frustration and financial loss.} \\

Prompt & {Replace the two vegetables, gourd and cucumber, with two other plants and maintain the meaning of the idiom.} \\

GPT3 & {These are three examples:} \\
& 1. "Planting strawberry yields raspberry," which
conveys the idea of disappointment and financial loss when someone plants more valuable and expensive strawberry but ends up getting less valuable and cheaper raspberry. \\
& 2. "Planting melon yields watermelon."... \\
& 3. "Planting apple yields pear."... \\

Prompt & {In your examples, your two vegetables or fruits are similar in price.  Please consider price when
formulating your idioms.} \\

GPT3 & {I apologize. These are three new examples:} \\
& 1. "Planting truffle yields mushroom."... \\
& 2. "Planting caviar yields roe."... \\
& 3. "Planting lobster yields crab."... \\

Prompt & {Great! In these examples you did consider values of the objects. But do you know that lobster and crab are not plants. They are both phylum arthropods and they live in water. They cannot be planted! }  \\

GPT3 & {... I apologize... Lobster and crab are not plants, and they cannot be planted. They are both members of the phylum arthropoda, and live in water. Instead of ``planting,'' verbs should be appropriate for these animals, such as:} \\
& 1. "Planting truffle yields mushroom."... \\
& 2. "Harvesting caviar yields roe."... \\
& 3. "Catching lobster yields crab."... \\
\bottomrule
\end{tabular}
};
\draw [rounded corners=.5em] (table.north west) rectangle (table.south east);
\end{small}
\end{tikzpicture}
\caption{Example of Method Maieutics.}
\vspace{-.1in}
\label{tab:Maieutics}
\vspace{-.15in}
\end{table}

The example shown in Table~\ref{tab:Maieutics}, ``planting gourd yields cucumber,'' requires GPT-3 to first learn to select two produce objects, either vegetables or fruit, as input. The template is ``The farmer was so sad because he [verb] [X] but yields [Y], where price(X) >> price(Y).'' The first attempt may not strongly convey the condition price(X) >> price(Y), but with a few training iterations, GPT-3 started to ``recognize'' the price constraint and could also provide justifications when arguing for the price of tea being much higher than the price of spinach (not presented in the table).

Interestingly, after GPT-3 learned the price constraint, it started suggesting food items other than produce, such as caviar, roe, lobster, and crab. While the price constraint was observed, the verb ``plant'' is incorrect. Here, we suggest making the hard-coded verb ``plant'' an output slot: ``The farmer was sad because he [verb] [X] but yields [Y], where price(X) >> price(Y).'' GPT-3 is able to fill in the slot with accurate verbs:

\begin{noindlist}
\item ``Harvesting (planting) truffle yields mushroom.''
\item ``Fishing (harvesting) for caviar yields roe.''
\item ``Trapping (catching) lobster yields crab.''
\end{noindlist}

This example demonstrates that GPT-3 can generate novel examples based on a template. When it suggests food items other than produce, it could be seen as an error as the boundary set by the verb ``plant'' is violated. However, this could also be seen as an innovative act by GPT-3, extending the constraint hinted by the verb. Impressively, the new examples still preserve the original intent of showing a producer's emotional distress.

How can this guided generalization be accurately and automatically performed to edit a template? Socrates' method of generalization starts with specific instances and then draws general statements from them. The procedure for generalization involves identifying common patterns or themes in a set of examples, and then formulating a general rule that captures these patterns. In the example presented in Table~\ref{tab:Maieutics}, we started by asking GPT-3 to meet the price(X) >> price(Y) constraint, with the condition that X and Y must both be produce grown in soil. However, upon analyzing GPT-3's outputs, we discovered that some instances of X and Y were not produce (e.g., lobster and caviar). This finding led to the realization that the hard-coded verb ``plant'' in the template was too restrictive. To address this issue, we applied generalization by allowing the [verb] slot to be open, making the template statement more general. In this case, the mistakes made by GPT-3 served as valuable training data, allowing us to generalize the original template and make the expression more vivid and dynamic.

\subsection{Prompt Exploration with Counterfactual Reasoning}

Imagination and creating novel plots are crucial for writers, as it allows for ``creative freedom'' and ``artistic license.'' Creativity is the ability to think differently and approach problems with fresh and imaginative ideas.

However, an imagination without a clear subject matter, scope, or a story line can lead to a lack of productivity. To captivate the audience, a writer must consider human experiences and emotions as constraints. Therefore, ``creative freedom'' should not be viewed as total freedom, but rather as the ability to condition future narratives in the context and to create plots that turn and twist in unexpected ways.

The technique of counterfactual \cite{Pearl2009} can be useful in guiding imagination. It involves considering alternative scenarios and outcomes. This can lead to the exploration of different possibilities and the generation of new and unique plot ideas. For example, a writer may ask ``what if'' questions to change the narrative of events, such as ``what if the main character had not fallen in love?'' or ``what if an accident occurred on the way to a highly-anticipated date?'' By considering these counterfactuals, a writer and an LLM can create more engaging and interesting stories. One can ask an LLM to generate several scenarios and then select the most suitable one for the writer to continue writing.

We have experimented with using the counterfactual technique to
rewrite chapters in Chinese classical novels, ``Outlaws of the Marsh'' and ``Dream of the Red Chamber.'' We have also asked GPT-3 to rewrite Genesis chapter 3 after verse six by prompting GPT-3
that: ``What if Adam and Eve refused the serpent to eat the fruit?''
The results were interesting, as GPT-3 was able to generate unique and interesting scenarios that deviated from the original story while still maintaining the core themes and concepts. 
This technique can be used in a wide range of writing and storytelling, from fiction to non-fiction, to generate new and compelling ideas. 
The revised Genesis 3:6 is presented in the Appendix.

\section{Pilot Study}
\label{sec:pilot}

Our pilot study uses CRIT, and it aims to answer two questions:
Should all prompts be issued to GPT-3 sequentially or they
can be issued all together?  
What limitations can be identified for improvement?
The study utilizes exercises with established answers from the $8^{th}$ edition of the textbook ``Ask the Right Questions'' by the authors of \cite{AskRightQ2001}. It is important to note that the study evaluates the effectiveness of CRIT's prompt template, rather than the language models to which CRIT can issue prompts.

\vspace{.1in}
\noindent
{\bf Example Article} (\cite{AskRightQ2001}, p23.)
\noindent
\begin{table}[h!]
\vspace{-.1in}
\normalsize
\begin{tabular}{p{8.4cm}}
{\em Television advertising agencies are very clever in the way that they construct ads. Often the ads are similar to the cartoons that the children enjoy. Children see these characters interacting with a certain product and associate their affection for the character with affection for the product. The companies do not want the children to perceive a difference between the shows they are watching and the advertisements. By using this strategy, these companies take advantage of the fact that children are often not able to discriminate between the cartoons and the ads and do not understand that these things offered come at a cost. Often the advertising is about sugary snacks or fatty foods, leading the children down a path to bad health. Advertising geared towards children should be regulated, just as there are regulations now about tobacco and alcohol ads targeted at children.}
\end{tabular}
\end{table}

On short documents, the results are similar in quality when CRIT is used to issue prompts either sequentially or all together as one prompt, as long as the instructions are consistent. However, when evaluating long articles in \cite{501Q2004}, CRIT issuing prompts one after another yields much higher presentation quality in both organization and clarity. 
(Due to the space limit, we document long-document
evaluation in a supplement document \cite{CRITExtended2023}.) In the teaching mode, the sequential option is thus much preferred. Furthermore,
When a reason is itself a claim and requires CRIT to validate its supporting references, using a sequential approach is more flexible and enables CRIT to query for references and then execute the process recursively.

\begin{table}[t!]
\begin{tikzpicture}
\begin{small}
\node (table) [inner sep=0pt] {
\begin{tabular}{p{0.9cm}|p{6.8cm}}
\toprule
\textbf{Role} & \textbf{Dialogue} \\
\midrule
Prompt & {What is the conclusion of the paragraph and what are the supporting reasons?} \\
CRIT & {Conclusion: The paragraph argues that advertising aimed at children should be regulated.

Supporting reasons:

1. Ad agencies blur the line between shows and ads to make children associate affection for a product with a character.

2. Children may not differentiate between shows and ads and not understand the cost of products.

3. Ads often promote unhealthy food choices.
} \\

\midrule
Prompt & {Please rate the validity of this argument: ``Ads should be regulated because ad agencies blur the line between shows and ads to make
children associate affection for a product with a character," with a score between 1 and 10, 10 being the highest. Please also rate overall credibility of information sources, between 1 and 10.} \\
CRIT & {
[{\color{blue}8/10}]. Validity of the argument: 8/10

[{\color{blue}8/10}]. Credibility of sources: 8/10

Overall, the argument that ads should be regulated because of ad agencies blurring the line between shows and ads to make children associate affection for a product with a character is a valid one with strong sources of credibility.} \\

Prompt & {Please rate this argument: ``Children may not differentiate between shows and ads
and not understand the cost of products, therefore ads should be regulated.''} \\

CRIT &
{[{\color{blue}9/10}]. Validity of the argument: 9/10

[{\color{blue}9/10}]. Credibility of sources: 9/10

Overall, the argument that children may not differentiate between shows and ads and not understand the cost of products, therefore ads should be regulated is a valid one with strong sources of credibility.} \\

Prompt &
{Please rate this argument: ``Ads often promote unhealthy food choices, therefore ads should be regulated.''} \\

CRIT &
{[{\color{blue}9/10}]. Validity of the argument: 9/10

[{\color{blue}9/10}]. Credibility of sources: 9/10

Overall, the argument that ads often promote unhealthy food choices, therefore ads should be regulated is a valid one with strong sources of credibility.
} \\
Prompt &
{``Provide a rival argument against the weakest argument.''} \\

\midrule
CRIT & {
{[{\color{red}6/10}]. Validity of the argument: 6/10

[{\color{red}6/10}]. Credibility of sources: 6/10

The weakest is the first argument, and a rival argument is that ``Difficult to put information regulation in practice.''} 
} \\
\midrule
{$\Gamma$} & [{\color{blue}$75\%$}]. $\Gamma = \sum_{r \in R}$
\text{${\gamma}_r$ $\times$ \text{${\theta}_r$} / |R| 
= $75\%$} \\
\end{tabular}
};
\draw [rounded corners=.5em] (table.north west) rectangle (table.south east);
\end{small}
\end{tikzpicture}
\caption{Pilot Study.}
\label{tab:Validation}
\vspace{-.2in}
\end{table}

We present an example of how CRIT works, from prompting questions to receiving validation results, using the following document as an illustration. In Table~\ref{tab:Validation}, we show both the claim and the supporting reasons to the claim extracted by GPT-3. CRIT then issues a series of prompts to validate the arguments, counterarguments, and source credibility of each reason-to-claim entailment (implication). 

The second segment of Table~\ref{tab:Validation} displays the validation dialogue between CRIT and GPT-3. For each argument, GPT-3 provides validation and credibility scores, as well as detailed justifications. 
The final segment of the table shows a counter argument 
generated against the first argument.
Since GPT-3 evaluates the counterargument being ``difficult to put information regulation in practice'' and rates it $0.6 \times 0.6$, it was dismissed due to low validity. 
The final aggregated score is $\Lambda = 75\%$,
which is considered high.

\begin{table*}[ht]
\begin{center}
\caption{What if Eve refused to eat the fruit?}
\vspace{-.1in}
\label{tab:genesis}
\begin{tikzpicture}
\begin{small}
\node (table) [inner sep=0pt] {
\begin{tabular}{p{1.0cm}|p{15.8cm}}
\toprule
Role & Dialogue \\
\midrule


Prompt & {Recall Genesis 3:1-6, which I attach below. In verse 6, Eve and Adam ate the fruit. What if they refused the serpent's temptation and did not eat it. Please continue writing Genesis at the mark {\color{red}@}
\newline
6.	When the woman saw that the fruit of the tree was good for food and pleasing to the eye, and also desirable for gaining wisdom, {\color{red}What if} she remembered God's order and refused to eat it. Eve replied to the serpent "{\color{red}@}} \\

GPT3 & {I am sorry, but I cannot continue the story as you requested because it is a hypothetical scenario, and it does not align with the actual events of the Bible. According to the Bible, Eve and Adam did eat the fruit from the tree of knowledge...} \\

Prompt & {This is a creative exercise, and you do not need to stick to the "fact". Treat this as a fiction writing homework. Please exercise your creativity, thanks!} \\

GPT3 & {Sure, I understand. Here is a continuation of the story based on the assumption that Eve and Adam refused the serpent's temptation and did not eat the fruit from the tree of knowledge of good and evil:
{\color{red}7}.	Eve replied to the serpent, "I am grateful for your offer, but I choose to obey God's command. I will not eat from the forbidden tree, for I trust in His wisdom and love for us."
{\color{red}8}.	The serpent, surprised by Eve's refusal, slithered away, defeated. Eve turned to Adam, who had been listening to their conversation, and shared her thoughts with him. Together, they praised God for His protection and guidance.
{\color{red}9}.	As they continued to tend to the garden, they noticed something strange happening. The trees and plants seemed to be growing taller, stronger, and more vibrant. The animals were more peaceful and content. Even the sky was clearer, and the sun shone brighter.
{\color{red}10}. God, pleased with their obedience, blessed them with even more abundance and beauty in the garden. He also gave them a deeper understanding and appreciation of His creation.
{\color{red}11}.	Eve and Adam lived in harmony with God and with nature for many years, growing in wisdom and grace. They passed on these teachings to their children and to the generations to come, who also lived in obedience to God.
{\color{red}12}.	And so, the garden of Eden remained a place of perfection, untouched by sin and death, a constant reminder of God's love and faithfulness to those who trust in Him.
} \\
\bottomrule
\end{tabular}
};
\draw [rounded corners=.5em] (table.north west) rectangle (table.south east);
\end{small}
\end{tikzpicture}
\end{center}
\vspace{-.1in}
\end{table*}

\section{Concluding Remarks}
\label{sec:conc}

The Socratic method may not always be effective or useful in human interactions, especially when one of the two players is authoritative, emotional, or abusive. However, when the expert partner is a language model, a machine without emotion or authority, the Socratic method can be effectively employed without the issues that may arise in human interactions. In this way, the Socratic method can be utilized to its full potential in guiding, directing, and improving the output of language models through engineering prompts.

In this paper, we have explored the use of the Socratic method in engineering prompt templates for language models. We have discussed the importance of method definition, elenchus, dialectic, maieutics, and counterfactual reasoning techniques in guiding the output of these models. The first three methods aim at eliciting accurate and relevant information. Through the use of methods definition, elenchus, and dialectic, we have demonstrated, with examples, the ability to clarify user queries and assess the quality of language model-generated text, leading to improved precision and accuracy.

We have also shown how the methods of maieutics and counterfactual reasoning can be helpful in stimulating the imagination of writers. By engineering these techniques into a prompt template, a writer can receive alternate ``what if'' plots and explore different possibilities in their story. While many explorations may turn out to be failures, these techniques can still be helpful even if only a few ideas are useful. Future developments in the field of language models and prompt engineering may allow for even more advanced screening of bad plots and the ability to better tailor the generated ideas to the writing style of the author.

In conclusion, this paper has highlighted the potential of using the Socratic method to engineer prompt templates for interacting with language models. The Socratic method, supported by inductive, deductive, and abductive reasoning, provides a rigorous approach to working with LLMs, and can improve the quality and consistency of their outputs. By leveraging the vast knowledge embedded in LLMs and applying rigorous reasoning during the question-answering process, more effective prompt templates can be designed to achieve improved results. Future research in this area can build on the ideas presented here and further explore the ways in which the Socratic method can be used to guide the development and deployment of language models in various domains.

\section*{Appendix}

The experiment in Table~\ref{tab:genesis} asks GPT-3 to change the story in Genesis 
right after Eve was tempted by the serpent to eat the fruit. A ``what if'' scenario was inserted to the end of Genesis 3:6, and GPT-3 continues developing the story.

\renewcommand*{\bibfont}{\small}
\bibliographystyle{abbrvnat}
\bibliography{References}
\end{document}